# Predicting stock prices with ChatGPT-annotated Reddit sentiment: Hype or reality?

Mateusz Kmak, Kamil Chmurzyński, Kamil Matejuk, Paweł Kotzbach, and Jan Kocoń

Department of Artificial Intelligence
Wrocław University of Science and Technology, Wrocław, Poland
`jan.kocon@pwr.edu.pl`

**Abstract.** The surge of retail investor activity on social media, exemplified by the 2021 GameStop short squeeze, raised questions about the influence of online sentiment on stock prices. This paper explores whether sentiment derived from social media discussions can meaningfully predict stock market movements. We focus on Reddit's r/wallstreetbets and analyze sentiment related to two companies: GameStop (GME) and AMC Entertainment (AMC). To assess sentiment's role, we employ two existing text-based sentiment analysis methods and introduce a third, a ChatGPT-annotated and fine-tuned RoBERTa-based model designed to better interpret the informal language and emojis prevalent in social media discussions. We use correlation and causality metrics to determine these models' predictive power. Surprisingly, our findings suggest that social media sentiment has only a weak correlation with stock prices. At the same time, simpler metrics, such as the volume of comments and Google search trends, exhibit stronger predictive signals. These results highlight the complexity of retail investor behavior and suggest that traditional sentiment analysis may not fully capture the nuances of market-moving online discussions.

**Keywords:** ChatGPT · Stock market · Social media · Sentiment

## 1 Introduction

The stock market is usually assumed to be largely controlled by large investment funds. Historically, small retail investors did not have enough funds to have a measurable impact on the price of stocks individually. However, the recent rise of social media allows for large-scale cooperation between previously unconnected individuals. This resulted in the situation at the beginning of 2021 when a financial analyst named Keith Gill (usually known under the nickname of "DeepFuckingValue") created a post on Reddit,in which he outlined his arguments for investing in Gamestop.

At the time, investing in Gamestop was largely assumed to be unprofitable, as the company's stock was at an all-time low, and some funds were trying to decrease the price further. Despite that, Keith Gill's arguments were accepted by a large amount of Reddit users. The stock started being bought by many of the platform's users, and the price started to increase. The movement began to spiral out of control, mainstream news platforms started covering it, and the whole situation eventually resulted in a $\sim 7900\%$ price increase of the stock.

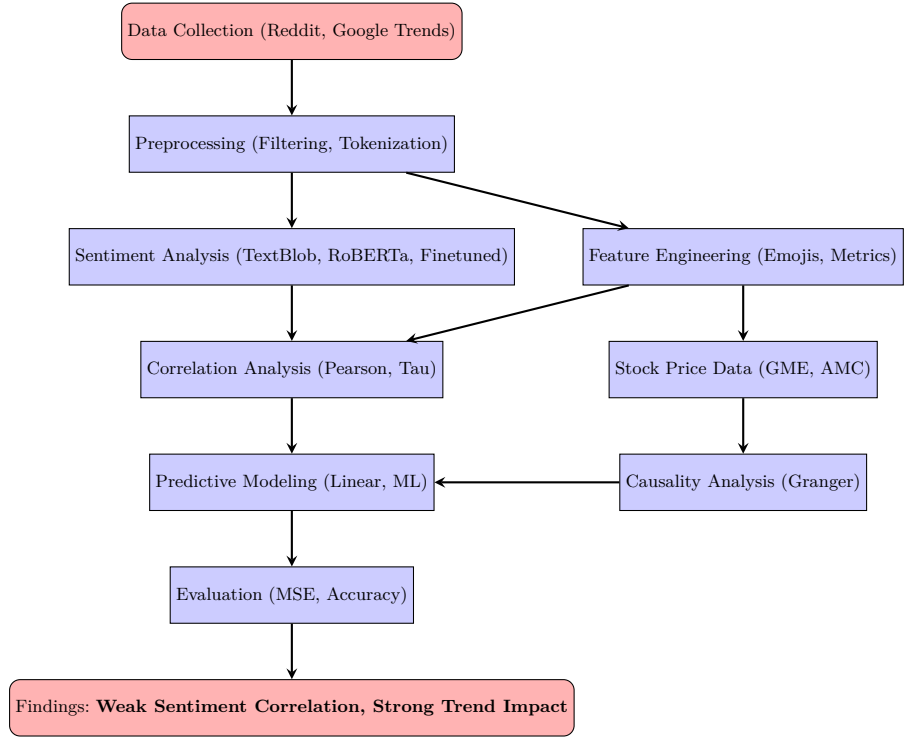

Fig. 1: System overview of sentiment and stock market analysis

This case inspired us to verify if there are other cases where one can attribute the change in stock price to social media. We have identified two companies for which we tested this theory. These companies are Gamestop (GME) and AMC Entertainment Holdings Inc (AMC). For each company, we have gathered a dataset of Reddit posts/comments mentioning it and measured if the sentiment and volume of these posts impact the company's price. The metrics used are:

1. Pearson correlation coefficient [22] – denoted as $r$, is a statistical measure that evaluates the linear relationship between two variables. It ranges from -1 to 1, where 1 indicates a perfect positive linear correlation, -1 indicates a perfect negative linear correlation, and 0 signifies no linear correlation.
2. Kendall Tau correlation [17] – denoted as $\tau$, is a non-parametric measure used to assess the ordinal association between two measured quantities. It

    evaluates the strength and direction of a relationship between two variables by comparing the ranks of their data points. A value of 1 indicates a perfect agreement, -1 indicates a perfect disagreement, and 0 suggests no association.
3. Granger causality [4] – a statistical hypothesis test for determining if one time series can predict another. It does not test for true causality but rather whether one variable's past values can help predict another's future values. The p-value in Granger causality tests indicates the probability of observing the test statistics if the null hypothesis (that one time series does not help predict the other) is true. A small p-value (typically $\leq 0.05$) indicates strong evidence against the null hypothesis, suggesting that the predictor time series has predictive power on the target series.

As an additional input, we checked these metrics for the number of Google searches (interest value from Google Trends) of the company to see if it also impacts the stock price, see Figure 1.

While two out-of-the-box models for classifying the sentiment of text were used, a new one was also created by finetuning an existing model based on RoBERTa. The fine-tuning (similar to [15, 19, 21, 20, 9, 13, 18]) was done to classify better the sentiment of posts, which used phrases very specific to the Reddit platform and a substantial amount of emojis. We provided the training data for finetuning by labeling the collected Reddit posts automatically with the help of ChatGPT, similar to [16, 14, 27, 12].

## 2   Related work

Since the GameStop black swan event, there has been a tremendous amount of work on this topic. There is a lot of insight into the social behaviors that drove people into doing so [3, 5, 8, 1], which offer explanations as to why the search for correlation and especially causality makes sense. Many researchers were trying to see if there were differences between before the GME short squeeze happened and afterward [26], and there certainly has been a change in how seriously social media is seen in affecting the stock market [6]. Although, it seems impossible to predict stock prices with Reddit [24, 10, 2] or any other social media outlets[11][7][23]. Some research has gone into analyzing the correlation between stock prices and various social indicators[25]. This has led us to explore, refine, and hopefully improve upon these approaches by comparing various stocks with each other in terms of their correlation and causality scores.

## 3   Datasets

As part of our work, we have prepared a dataset of Reddit posts from the */r/wallstreetbets* subreddit, which had a significant role in the Gamestop situation. The creation dates of the posts range between 2021.01.04 and 2021.03.31. In total, about 7 million posts were collected.

### 3.1   Emoji-driven Nature of Dataset

Emojis were segmented as one of the biggest differences between the dataset of Reddit posts and the original financial jargon the base model was trained on. From the new dataset, 561563 posts contained at least one emoji, with 1349 unique emojis overall. The most frequent usages of emojis were:

- 🚀: Associated with the phrase *To the moon*, meaning the stock price will skyrocket. Appeared in the dataset 1445364 times in 253700 posts.
- 💎: Associated with the phrase *Diamond hands*, meaning the investor refrains from selling an investment despite downturns or losses, believing in the long-term gains. Appeared in the dataset 451634 times in 162433 posts.
- 🙌: Associated with the phrase *Diamond hands*, as above. Appeared in the dataset 188311 times in 67247 posts.
- 🦍: Associated with users of the subreddit *r/wallstreetbets* calling themselves apes, meaning social-media traders battling against powerful institutional investors. Appeared in the dataset 113211 times in 41395 posts.

### 3.2   ChatGPT Annotation

Reddit posts containing emojis were fed in batches of 100 into ChatGPT. Below is an example classification prompt and the system's response.

---

**Task: Data annotation using ChatGPT**

**Prompt**

For each line of text below, separated by —, specify sentiment towards the stock as positive, neutral, or negative.
The context is meme stocks.
Respond with sentiment for each text in a separate line, with corresponding id.
Don't add any decorators, just lowercase response in one word.
Make sure to give the response for each id.
Here are the texts:
—
1: To the Muricans 🇺🇸 getting their stimmys today. Don't do the usual Murican thing and shit the bed. Pls and thank you. From a UK Ape 🦍 ✌️
—
2: 🙌🙌🙌🙌
—
3: THIS IS A VOLATILE FUCKING STOCK RIGHT NOW YOU GOTTA HAVE 💎🖐️
—
4. 😭😭😭😭😭😭😭😭😭😭 —
...
—
99: Fun game GME today! buy 🕛market 📈20 mins later..... 🏃💰💰....🛌💤
—
100: just got a fraction. im broke but it's what i can do 💎🙌
—

**ChatGPT answer**

1: positive
2: positive
3: positive
4: negative
...
99: positive
100: positive

### 3.3 ChatGPT Augumentation

As negative labels were very sparse in the collected dataset, ChatGPT was used again to generate negative examples based on negative samples from the dataset. Negative Reddit posts were fed in batches of 50 into ChatGPT, expecting to get 50 new samples generated in response. Below is an example generation prompt and the system's response.

---

**Task: Data generation using ChatGPT**

**Prompt**

Here are comments about Gamestock squeeze and meme stocks (stonks).
All of them were marked as negative.
Each one is divided with —.
Learn the style of these comments. Look at emojis.
Respond with 50 new comments, also negative, in the same style. Separate each generated comment with — and give it an index number starting from 1.
Here are the texts:
—
1: Access denied 🙅
—
2: GME IS DONE THE STOCK MARKET IS NEVER THAT EASY 🤣 AT LEAST YOU GUYS KNOW NOW 💥
—
3: Lol GME are losing enough money as it is without taking on another money losing operation 🤣🤣🤣
—
...
—
100: What an idiot we have here
—

**ChatGPT answer**

1: Classic move, blame everyone else for your failed predictions and cry foul. 🙅😂
2: Technical analysis skills as impressive as a toddler drawing with crayons. Get a grip, OP! 🙅
3: Denied access to sanity and logic, just like your stock predictions. 🙅📈
4: The only thing dropping faster than the price is the IQ of those defending the plunge. 📉🤣
...
50: Another stonk expert realizing the market isn't a playground for easy gains. 🤣💥💤

---

Considering the sizes of each class and time and resource constraints, 10525 samples were used for training: 7010 positive, 2219 neutral, and 1296 negative.

## 4 Models

### 4.1 TextBlob

**TextBlob** stands out in the field of natural language processing (NLP) as a versatile and user-friendly Python library. It is built upon the foundations of other well-established Python libraries like NLTK and Pattern and provides a comprehensive toolkit for textual data analysis.

The sentiment analysis feature of TextBlob is particularly noteworthy. It analyzes the emotional undertone of textual data, categorizing it as positive, negative, or neutral. The value we analyze is *polarity*, which is a measure on a scale of -1 to 1, where -1 denotes a highly negative sentiment and 1 represents a highly positive sentiment. A score of 0 indicates neutrality.

### 4.2 Financial-RoBERTa

**Financial-RoBERTa** represents a significant advancement in Natural Language Processing (NLP) applied to financial texts. This model adapts the RoBERTa Large language model, pre-trained and fine-tuned to specialize in the sentiment analysis of various financial documents. The domains of documents it can analyze include, but are not limited to, Financial Statements, Earnings Announcements, Earnings Call Transcripts, Corporate Social Responsibility (CSR) Reports, Environmental, Social, Governance (ESG) News, and other pertinent Financial News.

The development of Financial-RoBERTa involved an extensive training process using a diverse and comprehensive corpus. This corpus comprises various financial documents, such as 10-K, 10-Q, and 8-K filings, Earnings Call Transcripts, CSR Reports, ESG News, and Financial News texts. By utilizing such a varied and relevant dataset, the model is finely attuned to the nuances and specificities of financial language and terminology.

### 4.3 Finetuned version of Financial-RoBERTa

The Financial-RoBERTa model, while robust in analyzing formal financial texts, initially faced challenges in interpreting colloquial language, particularly prevalent in social media contexts. To address this, a specialized version of Financial-RoBERTa has been developed and fine-tuned using a unique dataset of Reddit posts. These posts, characterized by informal language, abundant use of emojis, and community-specific jargon, provide a rich resource for adapting the model to a more diverse linguistic environment.

The fine-tuning process involved the integration of a large corpus of Reddit posts, each labeled with sentiment tags generated by ChatGPT. This approach leveraged ChatGPT's advanced understanding of colloquial expressions and emoji usage to accurately tag each post with a corresponding sentiment label: Positive, Negative, or Neutral. The inclusion of this colloquially-rich and emoji-laden dataset aimed to enhance the Financial-RoBERTa model's proficiency in handling non-traditional financial texts, which are becoming increasingly relevant in today's digital and social-media-driven financial landscape.

Originally, Financial-RoBERTa's training on formal financial documents limited its effectiveness in contexts where informal language prevailed. This gap was particularly noticeable when analyzing texts from platforms like Reddit, where users frequently employ a casual tone, slang, and visual elements like emojis to express sentiments. By incorporating this informal, emoji-rich dataset into the fine-tuning process, the enhanced Financial-RoBERTa model now demonstrates improved accuracy and adaptability in analyzing texts that deviate from conventional financial language.

The fine-tuned version of the Financial-RoBERTa model retains its original proficiency in analyzing formal financial texts while extending its capabilities to comprehend and accurately classify sentiments in less formal, emoji-enriched texts. This advancement makes the model an even more versatile tool for financial analysts who seek to harness insights from a broader range of sources, including social media platforms where investors and consumers frequently express opinions and reactions colloquially.

## 5 Experiments

For every previously mentioned company, we have performed the following experiments:

1. Calculating stock price correlation and causality metrics for the daily number of comments and Google Trends values (0-100).
2. Calculating stock price correlation and causality metrics for the sentiment produced by TextBlob, Financial-RoBERTa, and the finetuned version of Financial-RoBERTa. Furthermore, as the finetuned version was supposed to handle emojis better, we also provided a baseline model, which counts the number of emojis in a text. We shall refer to this model as the Emoji Counter.

When we calculate correlation and causality metrics, we also provide them for two modifications of the original stock price data:

1. *SHIFTED* - we shift the stock price data by one day to the past. That way, we calculate the metrics as if we're trying to predict the price of the current day, not the next day.
2. *STATIONARY* - we operate not on absolute stock price values but on the daily price change. We believe this type of information is more suited for the Granger causality test, as it works best for stationary data.
3. *SHIFTED STATIONARY* - both STATIONARY and SHIFTED modifications are applied.

## 6 Results

### 6.1 Gamestop

In Table 1, a significant correlation is observed between stock price and the number of comments, indicating that social media discussions align closely with market movements, Fig. 2. The correlation with Google Trends values is slightly weaker, suggesting that while public interest follows price changes, it may not be as immediate as social media reactions. This difference implies that active trader discussions have a stronger short-term impact on stock fluctuations, whereas search trends may reflect broader, less engaged curiosity. Future research could further explore how different online activities influence stock behavior across various market conditions.

In Table 2, the stock price shows symmetrical causal relationships with the number of comments and Google Trends values, particularly after applying the

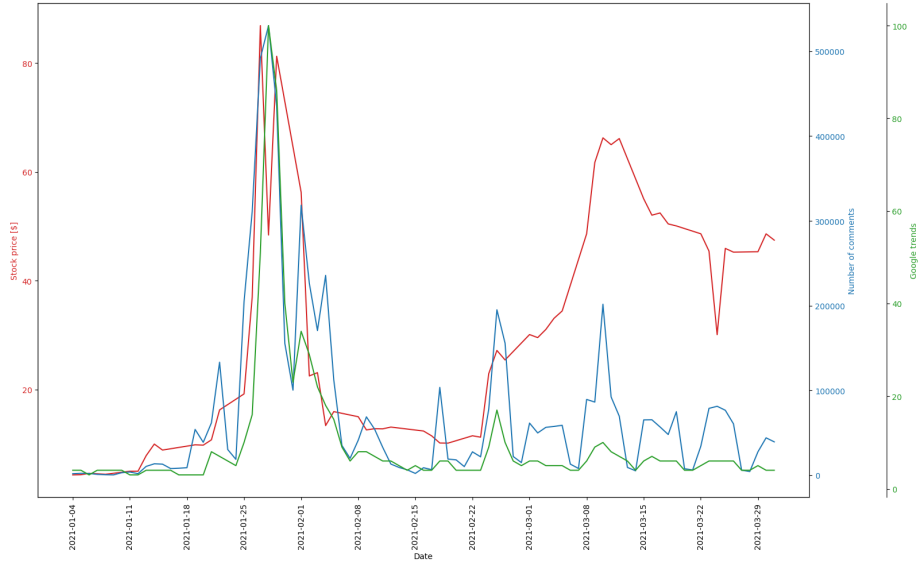

Fig. 2: Stock price, number of comments, and Google Trends value for Gamestop

| Type | Shifted | Value | Correlation |
|---|---|---|---|
| Pearson | false | Number of comments | **0.521863** |
| Pearson | true | Number of comments | 0.470033 |
| Kendall/Tau | false | Number of comments | 0.479781 |
| Kendall/Tau | true | Number of comments | 0.442623 |
| Pearson | false | Google Trends | 0.426960 |
| Pearson | true | Google Trends | 0.434461 |
| Kendall/Tau | false | Google Trends | 0.410597 |
| Kendall/Tau | true | Google Trends | 0.407103 |

Table 1: Stock price correlation measurements for Gamestop

| Cause | Effect | p | Stationary | Shifted |
|---|---|---|---|---|
| Stock price | Number of comments | 0.9116 | false | false |
| Stock price | Number of comments | **0.0408** | true | false |
| Number of comments | Stock price | 0.4686 | false | false |
| Number of comments | Stock price | **0.0180** | true | false |
| Number of comments | Stock price | **0.0406** | false | true |
| Number of comments | Stock price | **0.0003** | true | true |
| Stock price | Google Trends | 0.2495 | false | false |
| Stock price | Google Trends | **0.0302** | true | false |
| Google Trends | Stock price | 0.9169 | false | false |
| Google Trends | Stock price | **0.0008** | true | false |
| Google Trends | Stock price | 0.3491 | false | true |
| Google Trends | Stock price | 0.0541 | true | true |

Table 2: Granger Causality measurements (p-value) for Gamestop

*STATIONARY* modification, according to the Granger Causality metric. This suggests a feedback loop where increased discussions and search trends can predict stock price movements while price fluctuations also drive online engagement. The bidirectional nature of this relationship supports the idea that modern markets are influenced by social sentiment alongside fundamental financial indicators. Stronger causal effects with stationarity adjustments indicate that investors react more to daily price changes than absolute stock levels, highlighting the role of event-driven volatility and momentum in shaping online discourse. Future research could examine whether specific social media interactions—such as sentiment, engagement, or trading discussions—impact stock movements more than general activity metrics.

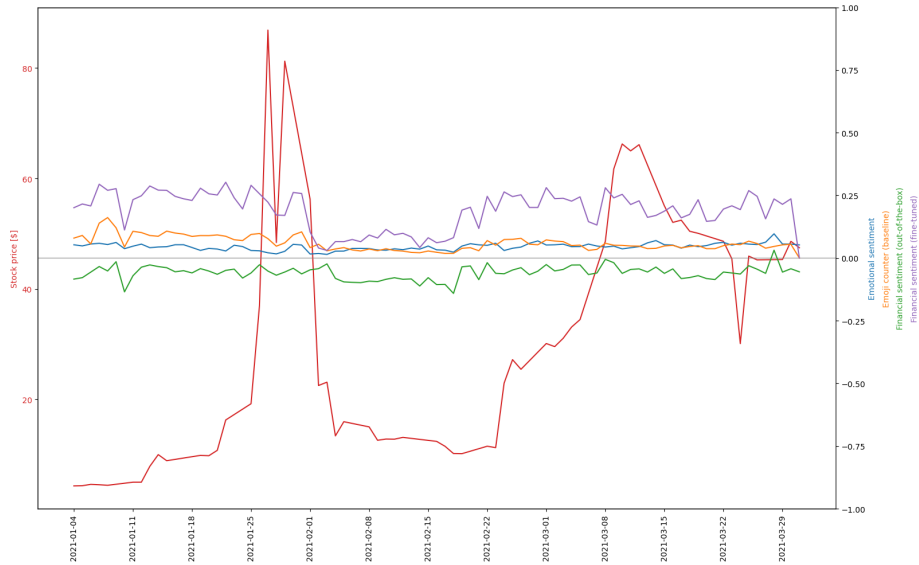

Fig. 3: Stock price and sentiment values for Gamestop

In Table 3, all sentiment values show very weak correlation with the stock price of Gamestop, indicating that traditional sentiment analysis may not effectively capture the factors influencing stock movements in this case, Fig. 3. In Table 4, some causal relationships were identified, suggesting an interaction between stock price movements and social media sentiment. Notably, the stock price can predict the number of emojis in Reddit posts about GME, indicating that price fluctuations influence how users express sentiment, particularly through informal and emotion-driven communication. Furthermore, the stock price has a symmetric causal relationship with the output of the fine-tuned financial model, suggesting that while sentiment analysis can reflect market trends, it may also be shaped by them. This bidirectional effect highlights the complex interplay between investor emotions, online discourse, and market dynamics, warranting further exploration into how different sentiment signals contribute to price movements.

| Type | Shifted | Sentiment | Correlation |
|---|---|---|---|
| Pearson | false | Emotional | -0.028362 |
| Pearson | true | Emotional | -0.058245 |
| Kendall/Tau | false | Emotional | -0.060109 |
| Kendall/Tau | true | Emotional | -0.040437 |
| Pearson | false | Financial (out-of-the-box) | 0.230485 |
| Pearson | true | Financial (out-of-the-box) | 0.135628 |
| Kendall/Tau | false | Financial (out-of-the-box) | 0.095082 |
| Kendall/Tau | true | Financial (out-of-the-box) | 0.029508 |
| Pearson | false | Emoji counter | -0.290720 |
| Pearson | true | Emoji counter | -0.393952 |
| Kendall/Tau | false | Emoji counter | -0.239344 |
| Kendall/Tau | true | Emoji counter | -0.339891 |
| Pearson | false | Financial (finetuned) | 0.053431 |
| Pearson | true | Financial (finetuned) | -0.123969 |
| Kendall/Tau | false | Financial (finetuned) | -0.107104 |
| Kendall/Tau | true | Financial (finetuned) | -0.218579 |

Table 3: Stock price correlation measurements for Gamestop (sentiment data)

| Cause | Effect | p | Stationary | Shifted |
|---|---|---|---|---|
| Stock price | Emotional | 0.9259 | false | false |
| Stock price | Emotional | 0.1407 | true | false |
| Emotional | Stock price | 0.4594 | false | false |
| Emotional | Stock price | 0.6635 | true | false |
| Emotional | Stock price | 0.7703 | false | true |
| Emotional | Stock price | 0.3562 | true | true |
| Stock price | Financial (out-of-the-box) | 0.8971 | false | false |
| Stock price | Financial (out-of-the-box) | 0.1127 | true | false |
| Financial (out-of-the-box) | Stock price | 0.1998 | false | false |
| Financial (out-of-the-box) | Stock price | 0.5392 | true | false |
| Financial (out-of-the-box) | Stock price | 0.0927 | false | true |
| Financial (out-of-the-box) | Stock price | 0.7953 | true | true |
| Stock price | Emoji counter | **0.0267** | false | false |
| Stock price | Emoji counter | 0.1010 | true | false |
| Emoji counter | Stock price | 0.5876 | false | false |
| Emoji counter | Stock price | 0.6167 | true | false |
| Emoji counter | Stock price | 0.4617 | false | true |
| Emoji counter | Stock price | 0.8832 | true | true |
| Stock price | Financial (finetuned) | **0.0097** | false | false |
| Stock price | Financial (finetuned) | 0.1816 | true | false |
| Financial (finetuned) | Stock price | 0.0550 | false | false |
| Financial (finetuned) | Stock price | 0.6349 | true | false |
| Financial (finetuned) | Stock price | **0.0120** | false | true |
| Financial (finetuned) | Stock price | 0.1488 | true | true |

Table 4: Granger Causality measurements (p-value) for Gamestop (sentiment data)

## 6.2 AMC

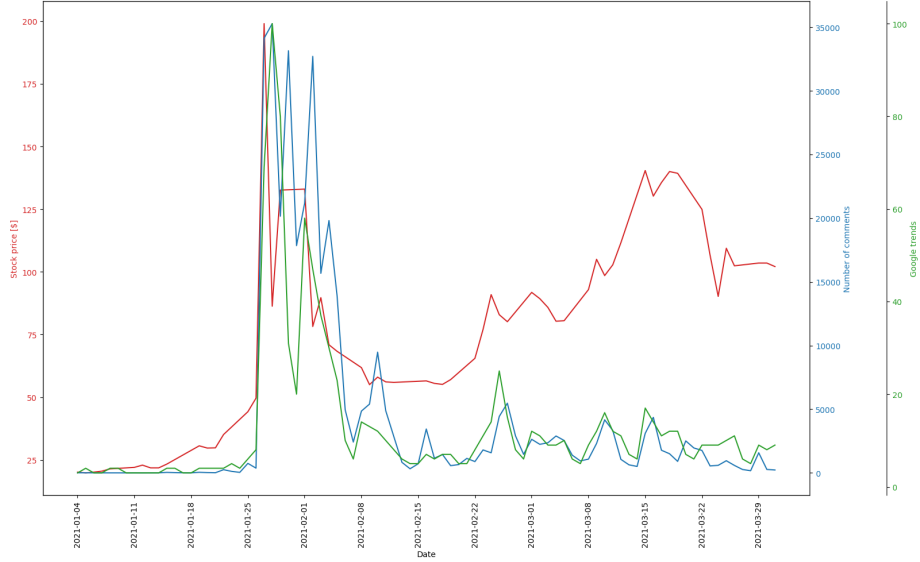

Fig. 4: Stock price, number of comments and Google Trends value for AMC

| Type | Shifted | Value | Correlation |
|---|---|---|---|
| Pearson | false | Number of comments | 0.377953 |
| Pearson | true | Number of comments | 0.387969 |
| Kendall/Tau | false | Number of comments | 0.363188 |
| Kendall/Tau | true | Number of comments | 0.363188 |
| Pearson | false | Google Trends | 0.475918 |
| Pearson | true | Google Trends | 0.482916 |
| Kendall/Tau | false | Google Trends | **0.551863** |
| Kendall/Tau | true | Google Trends | **0.537192** |

Table 5: Stock price correlation measurements for AMC

In Table 5, a significant correlation is observed between the stock price and Google Trends values, suggesting that public interest closely follows market movements, Fig. 4. The correlation with the number of comments is slightly weaker, indicating that while social media discussions align with price changes, they may not be as immediate or strong as search trends. This difference implies that Google search activity, often driven by broader awareness and media coverage, may serve as a stronger indicator of investor attention compared to direct engagement in online discussions.

| Cause | Effect | p | Stationary | Shifted |
|---|---|---|---|---|
| Stock price | Number of comments | 0.2567 | false | false |
| Stock price | Number of comments | **0.0007** | true | false |
| Number of comments | Stock price | 0.2459 | false | false |
| Number of comments | Stock price | 0.0823 | true | false |
| Number of comments | Stock price | 0.3028 | false | true |
| Number of comments | Stock price | **0.0000** | true | true |
| Stock price | Google Trends | 0.2541 | false | false |
| Stock price | Google Trends | **0.0060** | true | false |
| Google Trends | Stock price | 0.7648 | false | false |
| Google Trends | Stock price | 0.5194 | true | false |
| Google Trends | Stock price | 0.1554 | false | true |
| Google Trends | Stock price | **0.0000** | true | true |

Table 6: Granger Causality measurements (p-value) for AMC

In Table 6, the stock price exhibits strong symmetrical causal relationships with the number of comments and Google Trends values, particularly after applying the *STATIONARY* modification, as indicated by the Granger Causality metric. This suggests a dynamic feedback loop where stock price fluctuations influence both online discussions and search interest, while increased public attention may also contribute to market movements. The stronger causality with stationarity adjustments highlights that investors and social media users react more to daily price changes rather than absolute stock levels, reinforcing the role of short-term volatility in driving online engagement and market sentiment.

In Table 7, all sentiment values show a very weak correlation with the stock price of AMC, suggesting that sentiment analysis alone may not be a strong predictor of price movements. This indicates that while social media discussions reflect investor emotions, they do not necessarily translate into direct market impact. The weak correlation also implies that other factors, such as trading volume, external news, or institutional activity, may play a more dominant role in shaping AMC's stock price than sentiment-driven retail investor behavior.

Interestingly, Table 8 shows that only the number of emojis had a measurable causal effect on the stock price, suggesting that non-textual sentiment indicators may play a unique role in market dynamics. This finding implies that emoji usage, which often conveys strong emotional reactions concisely, could be a more immediate and reliable signal of investor sentiment compared to traditional text-based sentiment analysis. The absence of significant causality for other sentiment measures further highlights the complexity of social media-driven market behavior, warranting further investigation into how different forms of online expression influence stock price movements.

| Type | Shifted | Sentiment | Correlation |
|---|---|---|---|
| Pearson | false | Emotional | -0.203991 |
| Pearson | true | Emotional | -0.215998 |
| Kendall/Tau | false | Emotional | -0.016954 |
| Kendall/Tau | true | Emotional | -0.018048 |
| Pearson | false | Financial (out-of-the-box) | 0.075078 |
| Pearson | true | Financial (out-of-the-box) | 0.047219 |
| Kendall/Tau | false | Financial (out-of-the-box) | -0.028887 |
| Kendall/Tau | true | Financial (out-of-the-box) | -0.097962 |
| Pearson | false | Emoji counter | 0.443890 |
| Pearson | true | Emoji counter | 0.308529 |
| Kendall/Tau | false | Emoji counter | 0.398197 |
| Kendall/Tau | true | Emoji counter | 0.306439 |
| Pearson | false | Financial (finetuned) | 0.025702 |
| Pearson | true | Financial (finetuned) | -0.042367 |
| Kendall/Tau | false | Financial (finetuned) | -0.035623 |
| Kendall/Tau | true | Financial (finetuned) | -0.134539 |

Table 7: Stock price correlation measurements for AMC (sentiment data)

| Cause | Effect | p | Stationary | Shifted |
|---|---|---|---|---|
| Stock price | Emotional | 0.0676 | false | false |
| Stock price | Emotional | 0.9952 | true | false |
| Emotional | Stock price | 0.5122 | false | false |
| Emotional | Stock price | 0.8800 | true | false |
| Emotional | Stock price | 0.6406 | false | true |
| Emotional | Stock price | 0.9490 | true | true |
| Stock price | Financial (out-of-the-box) | 0.6116 | false | false |
| Stock price | Financial (out-of-the-box) | 0.8307 | true | false |
| Financial (out-of-the-box) | Stock price | 0.4661 | false | false |
| Financial (out-of-the-box) | Stock price | 0.6542 | true | false |
| Financial (out-of-the-box) | Stock price | 0.6501 | false | true |
| Financial (out-of-the-box) | Stock price | 0.8125 | true | true |
| Stock price | Emoji counter | 0.5455 | false | false |
| Stock price | Emoji counter | 0.0503 | true | false |
| Emoji counter | Stock price | **0.0073** | false | false |
| Emoji counter | Stock price | 0.1809 | true | false |
| Emoji counter | Stock price | **0.0092** | false | true |
| Emoji counter | Stock price | 0.7610 | true | true |
| Stock price | Financial (finetuned) | 0.8773 | false | false |
| Stock price | Financial (finetuned) | 0.6278 | true | false |
| Financial (finetuned) | Stock price | 0.5053 | false | false |
| Financial (finetuned) | Stock price | 0.7419 | true | false |
| Financial (finetuned) | Stock price | 0.4976 | false | true |
| Financial (finetuned) | Stock price | 0.9206 | true | true |

Table 8: Granger Causality measurements (p-value) for AMC (sentiment data)

## 7  Conclusions and Future Work

Overall, sentiment values were neither strongly correlated nor causally related to either company's stock price. The model finetuned by us supplied the best results, but they were still statistically insignificant. Surprisingly, simpler metrics like the number of comments or Google Trends values were meaningfully correlated. Furthermore, the Granger Causality values imply a strong causal relationship for these simple metrics, although the *stationary* modification is needed for that to be true. Interestingly, it seems that this causal relation is symmetric - the number of comments and Google searches impact the stock price change, and the stock price change impacts the number of comments and Google searches. More research is needed to determine if some other method of determining sentiment can be found to provide values with a causal relationship with the stock price. Furthermore, it should be checked if the findings in this paper can be replicated for different companies and periods.

## Acknowledgements

Financed by: (1) the National Science Centre, Poland (2021/41/B/ST6/04471, JK); (2) CLARIN ERIC (2024–2026), funded by the Polish Minister of Science (agreement no. 2024/WK/01); (3) CLARIN-PL, the European Regional Development Fund, FENG programme (FENG.02.04-IP.040004/24); (4) statutory funds of the Department of Artificial Intelligence, Wroclaw Tech; (5) the Polish Ministry of Education and Science ("International Projects Co-Funded" programme); (6) the European Union, Horizon Europe (grant no. 101086321, OMINO); (7) the EU project "DARIAH-PL", under investment A2.4.1 of the National Recovery and Resilience Plan. The views expressed are those of the authors and do not necessarily reflect those of the EU or the European Research Executive Agency.